\documentclass[letterpaper, 10 pt, conference]{ieeeconf}  
\usepackage{amsmath}
\usepackage{amssymb} 
\usepackage{graphicx}
\usepackage{float}
\usepackage{lipsum} 
\usepackage{placeins}
\usepackage{color}
\usepackage{bm}

\usepackage[style=ieee]{biblatex}

\DeclareSourcemap{
  \maps{
    \map{
      \pertype{misc}
      \step[fieldset=language, null]
      \step[fieldset=url, null]
      \step[fieldset=doi, null]
      \step[fieldset=issn, null]
      \step[fieldset=isbn, null]
      \step[fieldset=editor, null]
      \step[fieldset=urldate, null]
      \step[fieldset=file, null]
    }
  }
}
\DeclareSourcemap{
  \maps{
    \map{
      \pertype{article}
      \step[fieldset=language, null]
      \step[fieldset=url, null]
      \step[fieldset=doi, null]
      \step[fieldset=issn, null]
      \step[fieldset=isbn, null]
      \step[fieldset=note, null]
      \step[fieldset=editor, null]
      \step[fieldset=urldate, null]
      \step[fieldset=file, null]
    }
  }
}
\DeclareSourcemap{
  \maps{
    \map{
      \pertype{inproceedings}
      \step[fieldset=language, null]
      \step[fieldset=url, null]
      \step[fieldset=doi, null]
      \step[fieldset=issn, null]
      \step[fieldset=isbn, null]
      \step[fieldset=note, null]
      \step[fieldset=editor, null]
      \step[fieldset=urldate, null]
      \step[fieldset=file, null]
    }
  }
}
\DeclareSourcemap{
  \maps{
    \map{
      \pertype{incollection}
      \step[fieldset=language, null]
      \step[fieldset=url, null]
      \step[fieldset=doi, null]
      \step[fieldset=issn, null]
      \step[fieldset=isbn, null]
      \step[fieldset=note, null]
      \step[fieldset=editor, null]
      \step[fieldset=urldate, null]
      \step[fieldset=file, null]
    }
  }
}

\IEEEoverridecommandlockouts                              

\title{\LARGE \bf
 Thruster-Enhanced Locomotion: A Decoupled Model \\Predictive Control with Learned Contact Residuals
}
\author{Chenghao Wang$^{1}$, Alireza Ramezani$^{1*}$
\thanks{$^{1}$ The author is with Department of Electrical and Computer Engineering, Northeastern University, Boston, MA, USA  { wang.chengh, a.ramezani@northeastern.edu}}%
\thanks{$^{*}$ Corresponding author. Email: {a.ramezani@northeastern.edu}}%
}

\begin{document}

\maketitle
\thispagestyle{empty}
\pagestyle{empty}

\begin{abstract}

Husky Carbon, a robot developed by Northeastern University, serves as a research platform to explore unification of posture manipulation and thrust vectoring. Unlike conventional quadrupeds, its joint actuators and thrusters enable enhanced control authority, facilitating thruster-assisted narrow-path walking. While a unified Model Predictive Control (MPC) framework optimizing both ground reaction forces and thruster forces could theoretically address this control problem, its feasibility is limited by the low torque-control bandwidth of the system's lightweight actuators. To overcome this challenge, we propose a decoupled control architecture: a Raibert-type controller governs legged locomotion using position-based control, while an MPC regulates the thrusters augmented by learned Contact Residual Dynamics (CRD) to account for leg-ground impacts. This separation bypasses the torque-control rate bottleneck while retaining the thruster MPC to explicitly account for leg-ground impact dynamics through learned residuals. We validate this approach through both simulation and hardware experiments, showing that the decoupled control architecture with CRD performs more stable behavior in terms of push recovery and cat-like walking gait compared to the decoupled controller without CRD.

\end{abstract}

\section{INTRODUCTION}
Legged robots excel in navigating uneven terrain like rocks, foliage or stairs through adaptive limb-based locomotion, overcoming obstacles impractical for others\cite{bledt_mit_2018}. Aerial platforms prioritize speed and vertical mobility, bypassing ground barriers like waterways or fences but face endurance limits. By merging legged and aerial capabilities, robots harness terrain agility with rapid, obstacle-agnostic flight, creating a hybrid system where legged precision complements aerial reach for unprecedented versatility in dynamic environments. There are two primary design strategies for achieving such multimodal locomotion: additive integration\cite{kim_bipedal_2021}\cite{chae_ballu2_2021}\cite{pitroda_conjugate_2024}, which combines dedicated legged and aerial subsystems, or appendage repurposing\cite{sihite_multi-modal_2023}\cite{salagame_letter_2022}, as seen in platforms like Husky-$\beta$, where limbs are dynamically reconfigured to serve dual roles, such as legs transforming into thrusters for hybrid legged-aerial mobility.

The appendage repurposing mechanism reveals more interesting research directions. In legged mode, the propeller provides supplementary lateral control authority to assist in roll stabilization. Conversely, in aerial mode, the leg’s movable degree of freedom facilitates thruster vectoring, which enables more agile maneuvers. In this paper, we focus on the control problem arising in the first scenario, specifically the use of a sagittal thruster for assisted walking.

A unified MPC framework could be the promising path\cite{wang_quadratic_2025}\cite{lee_enhanced_2023}\cite{sihite_posture_2024}\cite{salagame_quadrupedal_2024} towards addressing the assisted walking problem, which takes both ground reaction force and thruster force as control inputs, and solves for the optimal solutions that achieve the desired state. However, its implementation is restrained by the lightweight actuators' low control rate for torque control\cite{hooks_implementation_2018}, which is critical for the unified Model Predictive Control (MPC) since the solution from the ground reaction could be mapped to joint space as torques. 
\begin{figure}[]
    \centering
    \includegraphics[width=\linewidth]{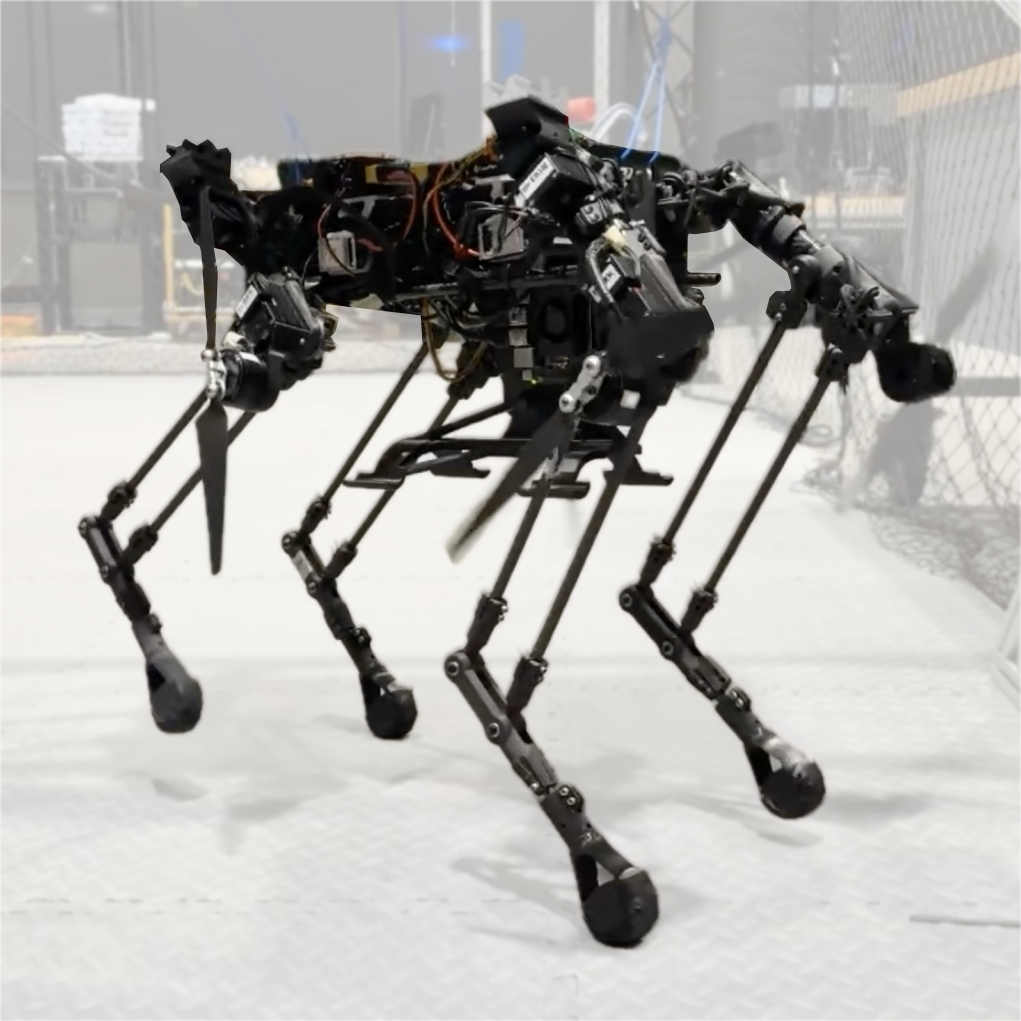} 
    \caption{Northeastern University Husky Carbon v.2 (or $\beta$ version) robot \cite{ramezani_generative_2021,sihite_multi-modal_2023,kim_bipedal_2021,pitroda_capture_2024,pitroda_enhanced_2024,pitroda_quadratic_2024,dangol_control_2021,dangol_feedback_2020,dangol_performance_2020,dangol_hzd-based_2021} executing a dynamic trotting gait with the sagittal thruster activated to enhance frontal stability during locomotion.}
    \label{fig:enter-label}
\end{figure}
To address this issue, we propose a decoupled control architecture comprising two main components: a legged controller and a thruster controller. The legged controller adopts a Raibert-type strategy to manage the swing and stance phases using position-based control, while the thruster controller employs MPC based on linearized angular dynamics, augmented with learned residual dynamics resulting from leg-ground impacts. 

The primary contributions of this work are:
\begin{itemize}
    \item Introducing a physics-informed loss function that enables the neural network to capture residual dynamics from leg-ground impacts through offline training on collected data.
    \item Designing a control pipeline for online execution that integrates a learned Contact Residual Dynamics (CRD) model into an MPC framework.
    \item Validating the approach through simulation and hardware experiments, demonstrating improved stability across two gaits, including a narrow-path cat gait.
\end{itemize}

\begin{figure}[!]
    \centering
    \includegraphics[width = \linewidth]{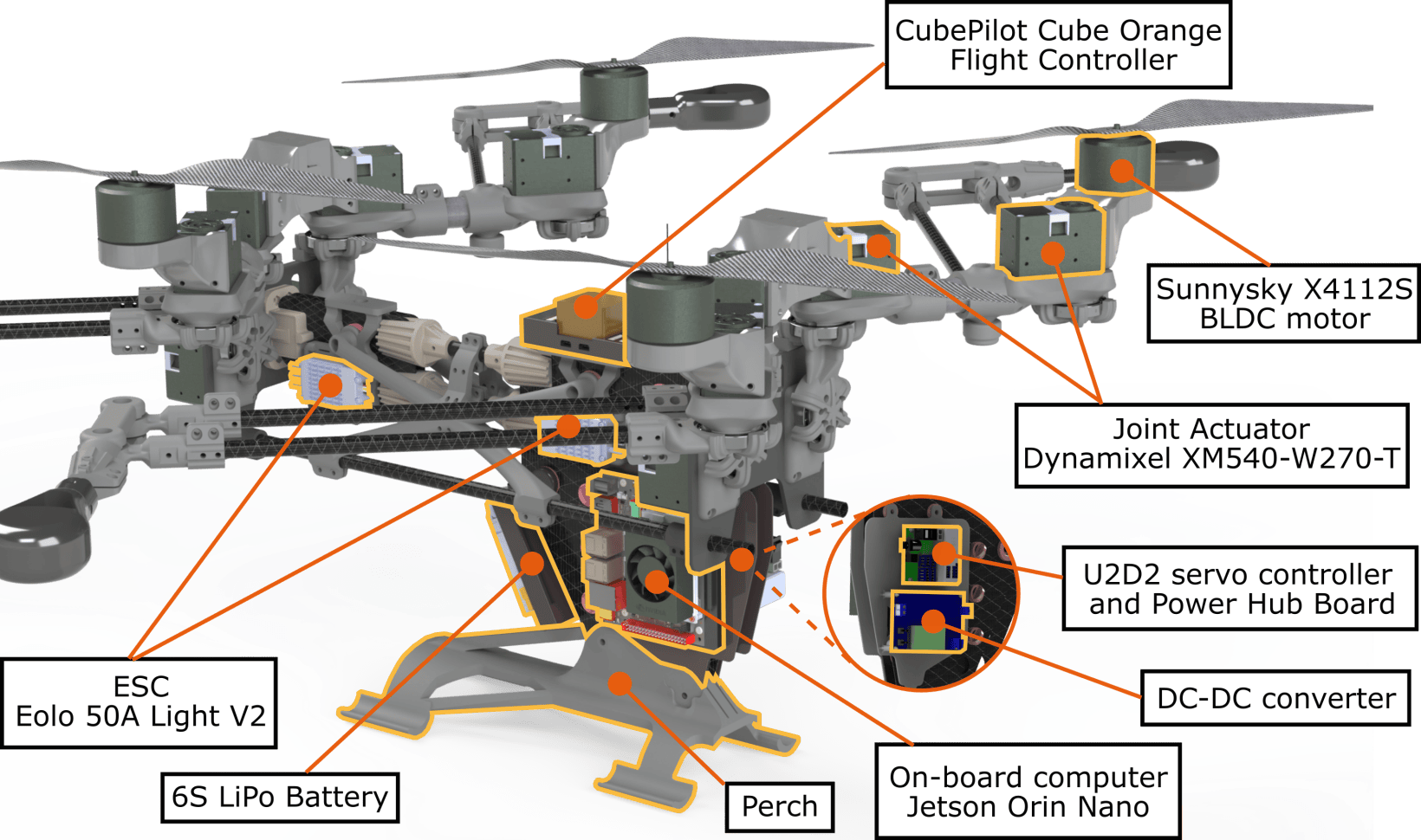}
    \caption{Hardware overview of Husky platform. }
    \label{fig:hw-overview}
\end{figure}

\section{Hardware Overview}
\label{sec:hw-overview}
Husky-$\beta$, shown in Fig.\ref{fig:hw-overview} is engineered for robust quadrupedal locomotion and aerial mobility through a bio-inspired morphing mechanism that enables seamless transitions between legged and UAV modes \cite{wang_leggedwalking_2023}. The robot features four BLDC-actuated propellers mounted at its knee joints, allowing it to generate upward thrust by extending a leg and orienting the propellers upward, like a conventional quadcopter. The primary design rationale for such a hybrid robot is balancing actuation power with overall weight, which is critical for achieving stable flight without sacrificing ground mobility.

To address the challenge, our design emphasizes several strategies. A lightweight structure is achieved by using materials such as 3D-printed Onyx polymer and carbon fiber for the main body and leg assemblies, which minimizes the robot's mass. High torque-to-weight actuators were selected; for example, the Dynamixel XH540-W270-T servo provides a stall torque of 10.6~Nm while weighing only 165~g, offering strong leg actuation with minimal weight penalty. Additionally, each leg is equipped with a compact propulsion unit—a combination of a SunnySky X4112S brushless motor, an EOLO 50A LIGHT ESC, and a 15\(\times\)5.5 double-blade propeller—that collectively generates up to 12~kg of thrust, yielding a thrust-to-weight ratio of approximately 1.8, ensuring sufficient thrust for stable flight.

\section{Method}
\label{sec:method}
\begin{figure}[!]
    \centering
    \includegraphics[width = \linewidth]{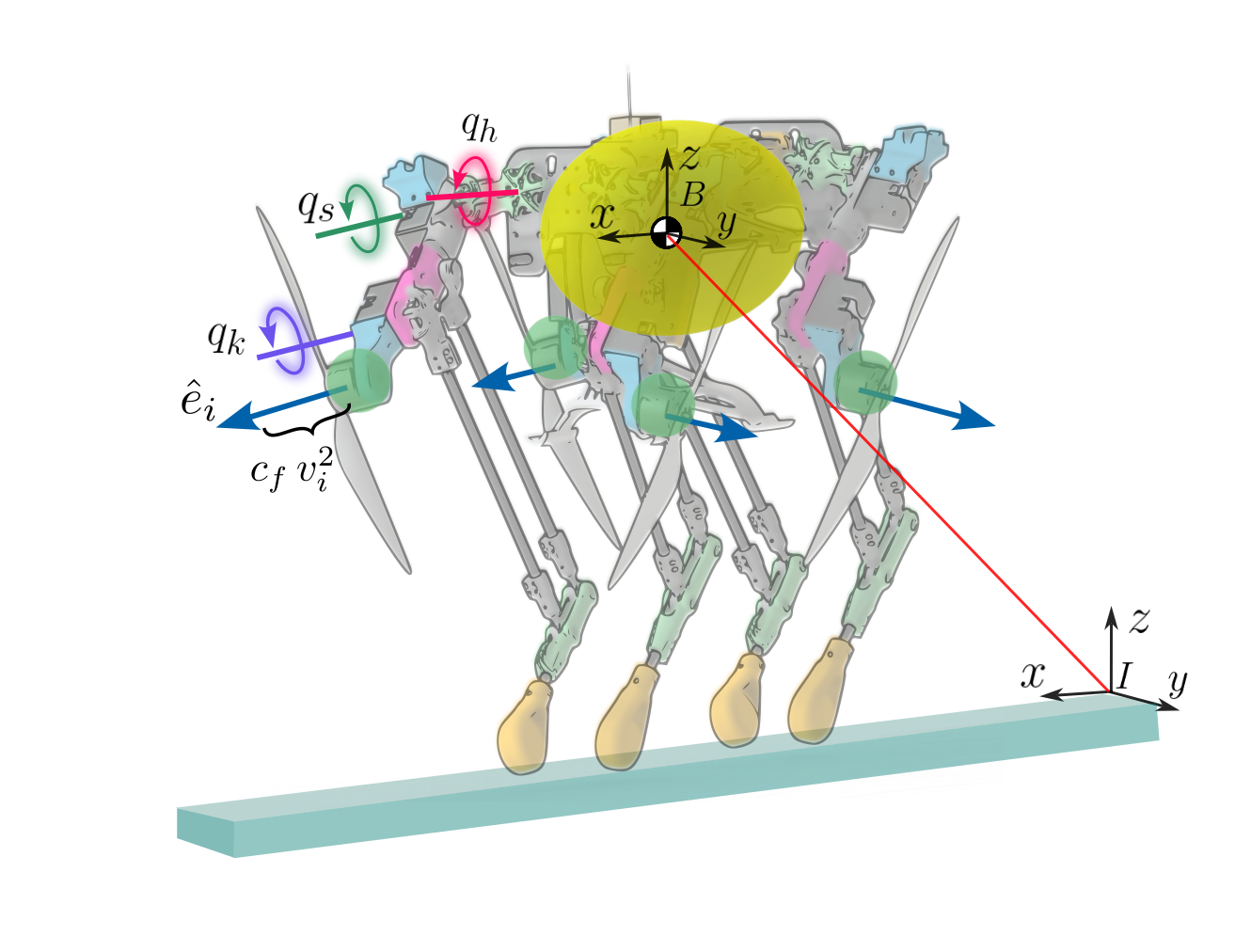}
    \caption{Free-body diagram of the Husky robot executing a narrow-path walking gait. The inertial frame $\{x, y, z\}_I$ is shown at the contact surface, while the body-fixed frame $\{x, y, z\}_B$ is located at the robot's CoM. Each leg includes a propeller providing a thrust force $f_i = c_f v_i^2$ along direction $\hat{\bm{e}}_i$, where $c_f$ is a force coefficient and $v_i$ is the propeller speed. Joint angles $q_h$, $q_s$, and $q_k$ represent the hip frontal, sagittal, and knee joints, respectively.}

    \label{fig:fbd-npw}
\end{figure}
\begin{figure*}[t]
    \centering
    \includegraphics[width=\linewidth]{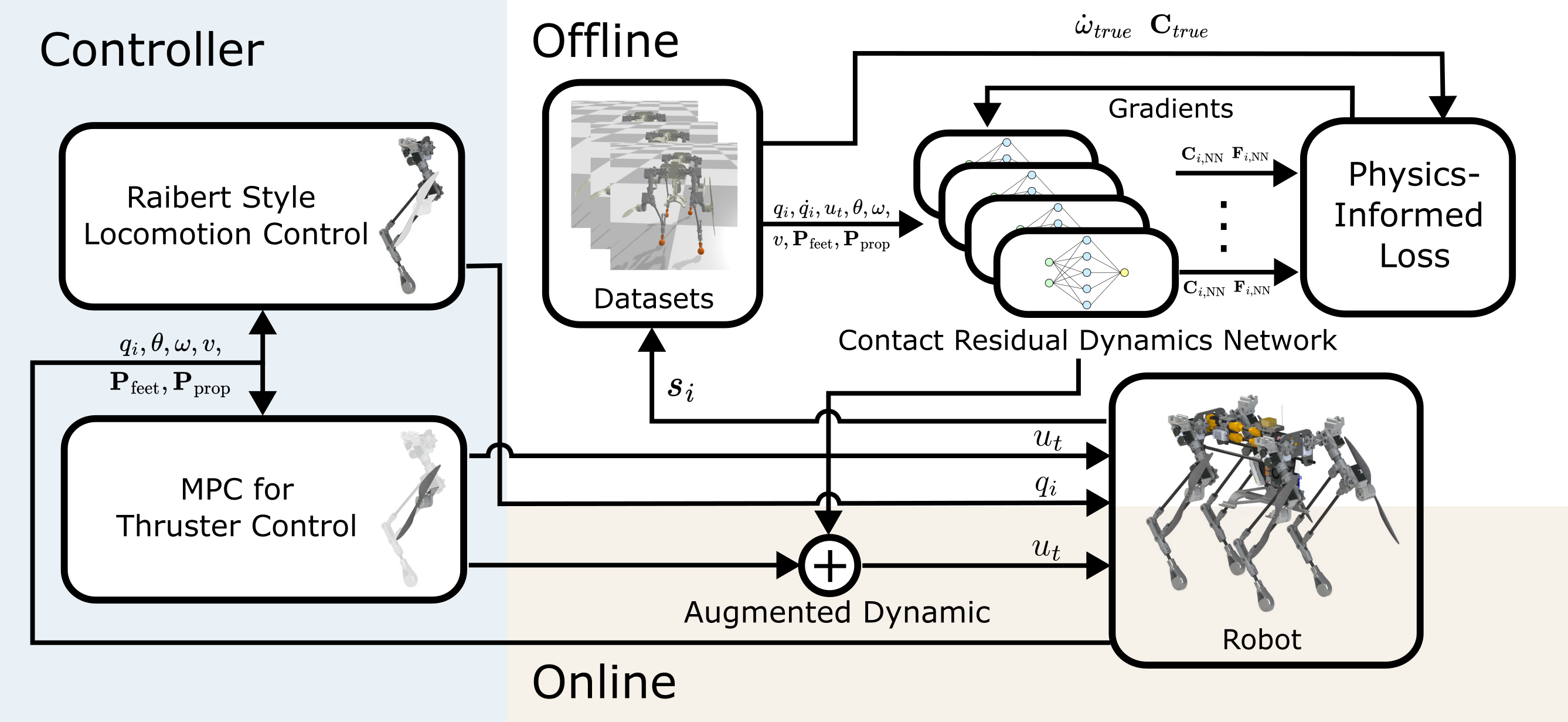}
    \caption{Overview of the proposed architecture: Decoupled controllers are used to collect a dataset, which is then employed to train a Contact Residual Dynamic (CRD) network composed of four identical sub-neural networks (one for each leg) using a physics-informed loss function. The trained network is then integrated with the nominal MPC to enable online execution. }
    \label{fig:hardware}
\end{figure*}

\subsection{Legged Control: Raibert Strategy}

For the swing phase, the desired foot placement is computed as:\\
\begin{equation}
\mathbf{p}_{d,i} = \mathbf{p}_{ref,i} + \frac{vT}{2} + k(v - v_d)
\end{equation}
where \(\mathbf{p}_{ref,i}\) is a reference position, \(v\) is the body velocity, \(T\) is the gait period, and \(v_d\) is the desired velocity.

During stance, the foot position is represented by
\begin{equation}
\mathbf{p}_i = \begin{bmatrix} x_i \\ y_i \end{bmatrix}
\end{equation}
and the vertical adjustment is computed using
\begin{equation}
\Delta z_i = \mathbf{p}_i^\top \mathbf{K}_{\text{ori}}\, \mathbf{e}
\end{equation}
with
\begin{equation}
\mathbf{K}_{\text{ori}} = \begin{bmatrix} k_{\text{pitch}} & 0 \\ 0 & k_{\text{roll}} \end{bmatrix}, \quad
\mathbf{e} = \begin{bmatrix} \text{pitch} \\ \text{roll} \end{bmatrix}
\end{equation}

A Jacobian-based method is employed to compute the robot's inverse kinematics using Pinocchio \cite{carpentier_pinocchio_2024}, which maps end-effector positions of the legs to joint positions.

\subsection{Thrust Control: Model Predictive Control (MPC)}

The state and control input for the thrust controller are defined as
\begin{equation}
\mathbf{x} = \begin{bmatrix} \boldsymbol{\theta} \\[0.5mm] \boldsymbol{\omega} \end{bmatrix} \in \mathbb{R}^6, \quad
\mathbf{u} = \begin{bmatrix} v_1 \\[0.5mm] v_2 \\[0.5mm] v_3 \\[0.5mm] v_4 \end{bmatrix}
\end{equation}
where \(\boldsymbol{\theta} \in \mathbb{R}^3 \) represents Euler angles for roll, pitch and yaw, $\boldsymbol{\omega} \in \mathbb{R}^3$ represents angular velocities for roll, pitch and yaw, \(v_i \in \mathbb{R}\) represents the propeller speeds which can be map to thruster force magnitude by
\begin{equation}
f_i = c_f\, v_i^2
\end{equation}
where \(c_f\) is the force constant.

The nominal angular dynamics, considering only the force contribution, are expressed as:
\begin{equation}
I\,\dot{\boldsymbol{\omega}}_{\text{nominal}} = \sum_{i=1}^4 \mathbf{r}_i \times \Bigl(c_f\, v_i^2\, \mathbf{\hat{e}}_i\Bigr)
\end{equation}
where \(\mathbf{r}_i\) is the vector from the center of mass to the \(i\)th propeller, and \(\mathbf{\hat{e}}_i\) is the unit vector in the thrust direction as  illustrated in Figure \ref{fig:fbd-npw}.

\subsection{Residual Dynamics Learning}
Unmodeled impulsive effects from leg-ground impacts introduce additional angular momentum changes, captured by a residual term:
\begin{equation}
\dot{\boldsymbol{\omega}}_{\text{true}} = \dot{\boldsymbol{\omega}}_{\text{nominal}} + \dot{\boldsymbol{\omega}}_{\text{residual}}
\end{equation}
with the residual approximated by\cite{di_carlo_dynamic_2018}
\begin{equation}
\dot{\boldsymbol{\omega}}_{\text{residual}} \approx I^{-1}\sum_{i=1}^4 \Bigl(\mathbf{d}_i \times \mathbf{F}_i\Bigr)
\end{equation}
where \(\mathbf{d}_i\) is the vector from the center of mass to the \(i\)th leg, and $\mathbf{F}_i$ is ground reaction force vector.

To enhance prediction accuracy, we employ a neural network to estimate the ground reaction forces and contact flags. Specifically, a \emph{Contact Residual Dynamics (CRD) Network} is replicated for each of the four legs. For each leg the network outputs both a 3D force prediction, \(\mathbf{F}_{i,\text{NN}}\), and a scalar contact probability, \(C_{i,\text{NN}}\). The estimated CRD can then be calculated by
\begin{equation}
\dot{\boldsymbol{\omega}}_{\text{residual}} = I^{-1}\sum_{i=1}^4 C_{i,\text{NN}} \Bigl(\mathbf{d}_i \times \mathbf{F}_{i,\text{NN}}\Bigr)
\label{eq-residual-dynamic}
\end{equation}

\subsubsection{Physics-informed loss function}

The network is trained using a composite loss that combines a physics-informed force loss and a contact prediction loss. The force loss is defined as the mean squared error (MSE) between the predicted and ground truth residual angular accelerations:
\begin{equation}
L_{\text{GRF}}= \Biggl\| I^{-1}\sum_{i=1}^4 C_{i,\text{NN}}\left( \mathbf{d}_i \times \mathbf{F}_{i,\text{NN}} \right) - (\dot{\boldsymbol{\omega}}_{\text{true}} - \dot{\boldsymbol{\omega}}_{\text{nominal}} )\Biggr\|^2
\end{equation}

The contact loss is formulated using binary cross-entropy:
\begin{equation}
\begin{aligned}
L_{\text{contact}} 
&= -\frac{1}{4} \sum_{i=1}^{4} \Bigl[\,
    C_{i,\text{GT}} \,\log\bigl(C_{i,\text{NN}}\bigr) \\
&\qquad\quad + \bigl(1 - C_{i,\text{GT}}\bigr)\,\log\bigl(1 - C_{i,\text{NN}}\bigr)\Bigr]
\end{aligned}
\end{equation}
where \(C_{i,\text{GT}} \in \{0,1\}\) is the ground truth contact flag and \(C_{i,\text{NN}}\) is the predicted contact probability for leg \(i\).

The overall training loss is a weighted sum of the two terms:
\begin{equation}
L = (1-\alpha)\, L_{\text{GRF}} + \alpha\, L_{\text{contact}}
\end{equation}
with the weighting parameter \(\alpha \in [0,1]\) controlling the trade-off between force estimation and contact prediction.

\subsubsection{Network architecture}
We employ a neural network with four identical subnetworks, one for each leg of the robot, to learn CRD. Each subnetwork processes a 21-dimensional feature vector (including the leg's joint angles and velocities, foot and propeller positions, the base's orientation, angular rates, linear velocity and thruster actions) through an input layer, four hidden layers with dimensions of 64, 128, 512, and 64 using ReLU as activation, and two outputs: a 3-unit force vector \(\mathbf{F}_{i,\text{NN}}\) and a 1-unit contact flag \(C_{i,\text{NN}}\). The sigmoid activation on the contact output restricts its value between 0 and 1, representing a binary contact state.

Because contact force and contact are closely correlated, both outputs share a common final embedding layer. The effective force is then computed as \(\mathbf{F}_{i,\text{NN}}\cdot C_{i,\text{NN}}\).

\subsection{CRD augmented MPC Formulation}

The continuous-time model, which integrates both nominal thrust dynamics and learned residual dynamics, is given by:
\begin{equation}
\begin{aligned}
\dot{\boldsymbol{\theta}} &= \boldsymbol{\omega}, \\
\dot{\boldsymbol{\omega}} &= I^{-1}\sum_{i=1}^4 \Biggl[ \mathbf{r}_i \times \Bigl(c_f\, v_i^2\, \mathbf{\hat{e}}_i\Bigr) + C_{i,\text{NN}}\, \Bigl(\mathbf{d}_i \times \mathbf{F}_{i,\text{NN}}\Bigr) \Biggr]
\end{aligned}
\end{equation}

\subsubsection{Discretization}
We first look at discretization for nominal dynamic. The nominal dynamics are linearized around an operating point, yielding a continuous-time state-space model:
\begin{equation}
    \dot{\mathbf{x}} = A_c \mathbf{x} + B_c \mathbf{u}
\end{equation}
Given a sampling period \(\Delta t\), the discrete-time state-space matrices are approximated as:
\begin{equation}
\begin{aligned}
    A_d &= \mathbb{I}_{6} + A_c \Delta t \\
    B_d &= B_c \Delta t
\end{aligned}
\end{equation}
where $\mathbb{I}_{6}$ is a identity matrix which have dimensions of $6 \times 6$. We assume that over a small \(\Delta t\), the state evolves linearly: \(\mathbf{x}_{k+1} \approx \mathbf{x}_k + \dot{\mathbf{x}}_k \Delta t\). Thus, the discrete-time model becomes:
\begin{equation}
    \mathbf{x}_{k+1} = A_d \mathbf{x}_k + B_d \mathbf{u}_k
\end{equation}

For an MPC horizon of \(H\) steps, the prediction model is constructed by iterating this equation. The state trajectory \(\mathbf{x} = [\mathbf{x}_1^\top, \mathbf{x}_2^\top, \ldots, \mathbf{x}_{H}^\top]^\top\) is expressed as:

\begin{equation}
\mathbf{x} = A_{\text{qp}} \mathbf{x}_0 + B_{\text{qp}} \mathbf{u}
\end{equation}
And then, we integrate the CRD network output into the discretized model. Specifically, for each horizon step \(k = 1, 2, \ldots, H\), the reference state is set as:

\begin{equation}
\mathbf{x}_{r,k} = \begin{bmatrix}
\boldsymbol{\theta}_0 + k \Delta t (\boldsymbol{\omega}_d + k \Delta t \dot{\boldsymbol{\omega}}_{\text{residual}}) \\
\boldsymbol{\omega}_d + k \Delta t \dot{\boldsymbol{\omega}}_{\text{residual}}
\end{bmatrix}
\end{equation}
where \(\dot{\boldsymbol{\omega}}_{\text{residual}}\) is obtained from CRD network output and calculated using Equation \ref{eq-residual-dynamic}.
\subsubsection{MPC formulation}
After integrating the learned residual dynamic into the discretized linear dynamic, the corresponding MPC problem is formulated as:

\begin{equation}
\begin{aligned}
\min_{\mathbf{u}} \quad & (\mathbf{x} - \mathbf{x}_r)^\top \bar{Q} (\mathbf{x} - \mathbf{x}_r) + \mathbf{u}^\top \bar{R} \mathbf{u}\\
\text{s.t.} \quad
& 0 < u_{t,i} < u_t^{\text{max}}, \quad i=1,\ldots,4
\end{aligned}
\end{equation}
where $\bar{Q}$ and $\bar{R}$ are block diagonal matrices composed of $Q$ and $R$, respectively.

\section{Experiment}

\subsection{Simulation Setup}
In the simulation experiment, a PyBullet simulator employing a high-fidelity model was used for both offline data collection and online execution. During the data collection phase, a series of randomized noise perturbations were introduced to enhance the robustness of the learned model. Specifically, random positional offsets were applied to the robot’s default positions—using uniform distributions to generate small variations in both the $x$ and $y$ directions. In addition, external forces were applied to the robot’s base along the x and y axes; these forces had randomized start times, durations, and magnitudes. Collectively, these noise injections allowed the dataset to capture a wide range of uncertainties, thus contributing to a more robust model.


In the online execution phase, we applied the same external perturbation to both the nominal-only MPC and the CRD-augmented MPC to validate the online effectiveness of our proposed method.
\subsection{Hardware Setup}
For the hardware experiment, we used hardware described in Section \ref{sec:hw-overview} to perform experiments in a tethered fashion to ensure safety. Two gait patterns were evaluated: one using a standard leg width configuration and another utilizing a narrower, cat-like gait. The narrow gait was specifically chosen to emphasize the CRD augmented controller's role in stabilizing roll dynamics under more challenging conditions. 

Due to the lack of contact sensors on the hardware, ground truth contact signals were unavailable for offline training. To address this limitation, we employed a transfer learning approach by training the hardware model using a pre-trained simulation-based CRD model. In this process, the weights of the layer responsible for outputting the contact probability were frozen. Since the force and contact outputs share the same latent embedding, this architecture inherently combines force feedback with contact estimation, compensating for the absence of direct contact sensing on the hardware.
\subsection{Results}
\subsubsection{Simulation Result}

\begin{figure}[!ht]
    \centering
    \includegraphics[width=\linewidth]{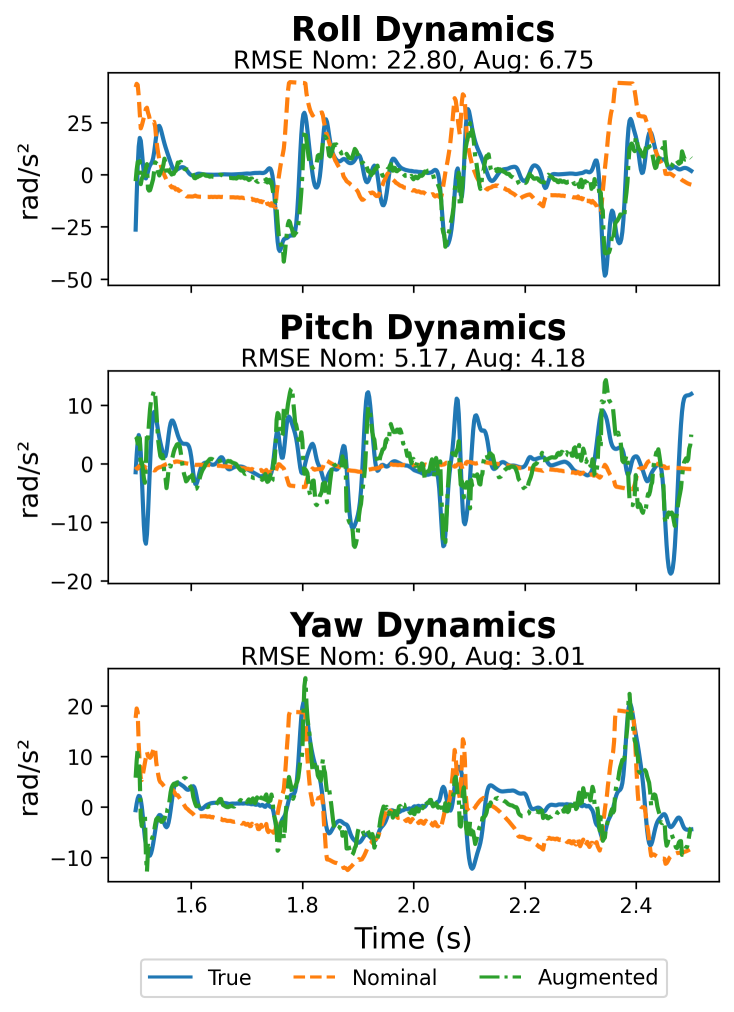}
    \caption{Angular dynamic $\dot{\boldsymbol{\omega}}$ comparison w and w/o CRD augmentation in simulation.}
    \label{fig:sim-result-offline-compare}
\end{figure}

Two simulation experiments were conducted. In the first experiment, a separate trajectory was collected offline, completely separate from the training data, to evaluate the performance of the proposed CRD network. During this evaluation (see Figure \ref{fig:sim-result-offline-compare}), the angular dynamics predicted by both the nominal MPC and the augmented MPC were compared against the ground truth data obtained directly from the PyBullet simulator. 

\begin{figure}[!ht]
    \centering\includegraphics[width=\linewidth]{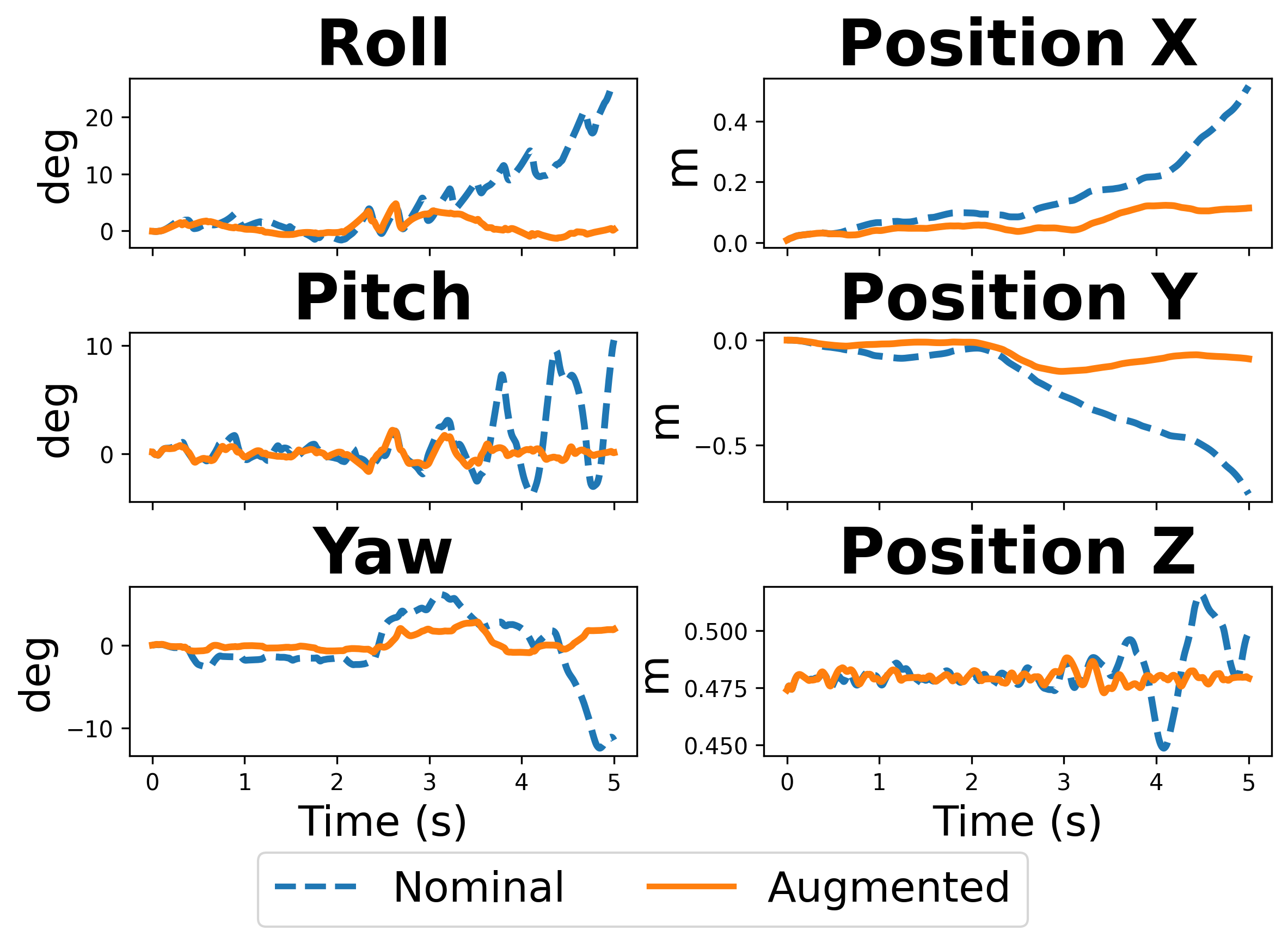}
    \caption{A 15N external force applied at 2s for 0.5s in simulation. With augmented dynamic recovered at 4s. Without augmented dynamic failed.}
    \label{fig:sim-result-push}
\end{figure}

The CRD-augmented dynamic model achieved a lower root mean square error (RMSE) compared to the nominal dynamic model across roll, pitch, and yaw. Notably, the most significant improvement was observed in the roll dynamics, due to the sagittal-facing thruster’s primary role in roll stabilization.
In the second experiment, following two seconds of trotting gait, a 15N push force with a 0.5-second duration was applied to the robot's y-axis for both controllers (see Figure \ref{fig:sim-result-push}). As expected both controllers exhibited roll instability; however, the CRD-augmented controller returned to a normal trotting gait at approximately 3.5 seconds, whereas the nominal controller failed to re-stabilize.
\subsubsection{Hardware Result}
\begin{figure*}[!ht]
    \centering
    \vspace{0.2cm}
    \includegraphics[width=\linewidth]{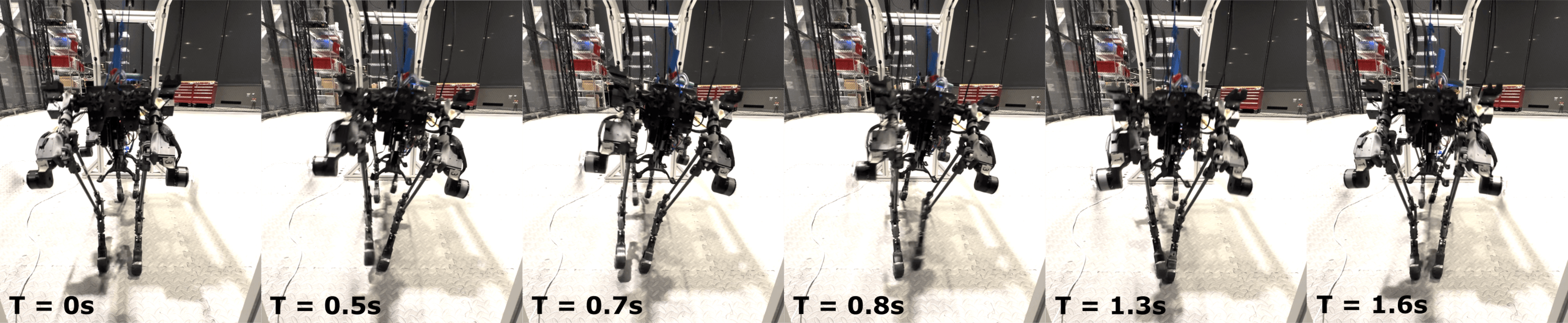}
    \caption{Roll stabilization utilize CRD augmented dynamic.}
    \label{fig:hw-result-npw-recovery}
\end{figure*}

To validate the effectiveness of the CRD-augmented dynamics, two gait types were employed in hardware experiments. First, a narrow-path cat gait at 1.5 Hz was utilized (see Figure \ref{fig:hw-result-npw-recovery}). Analysis of the time-series data reveals that, starting from a roll-stable configuration, the robot began rolling leftward at approximately 0.5 s, prompting the left propeller to activate in order to counteract the instability. 


\begin{figure}[!ht]
    \centering\includegraphics[width=0.95\linewidth]{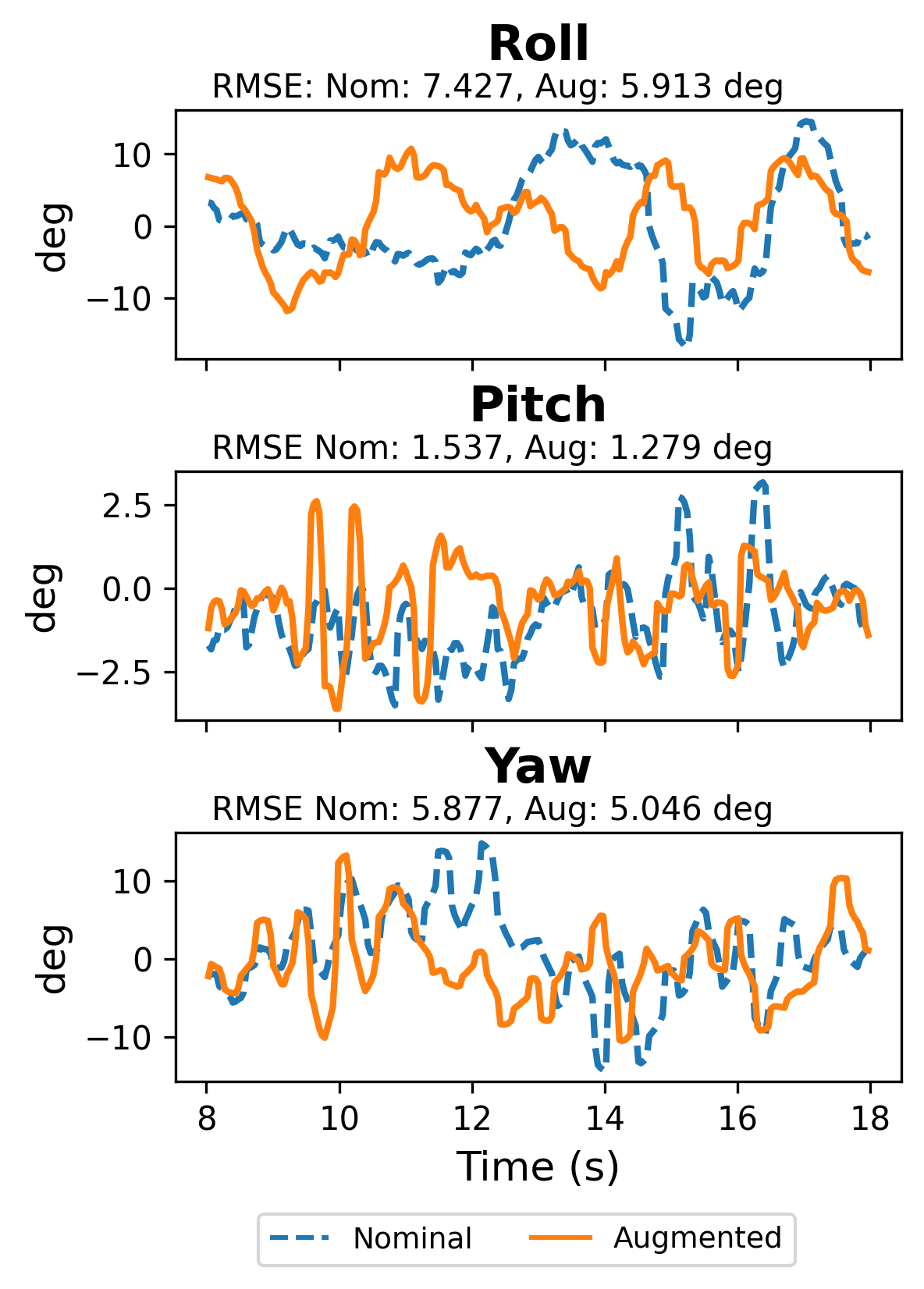}
    \caption{State comparison w and w/o augmentation under a cat-type walking gait on hardware.}
    \label{fig:hw-result-npw}
\end{figure}

Furthermore, a comparison between the CRD-augmented and nominal dynamic-based cat walking gait was performed (see Figure \ref{fig:hw-result-npw}). The results indicate that the CRD-augmented dynamic achieved a lower root mean square error (RMSE) across all three axes, roll, pitch, and yaw when compared to the nominal dynamic. The most significant improvement was observed in the roll dynamics, due to the side-mounted thruster that predominantly contributes to roll stabilization.

\begin{figure*}[!ht]
    \centering
    \includegraphics[width=\linewidth]{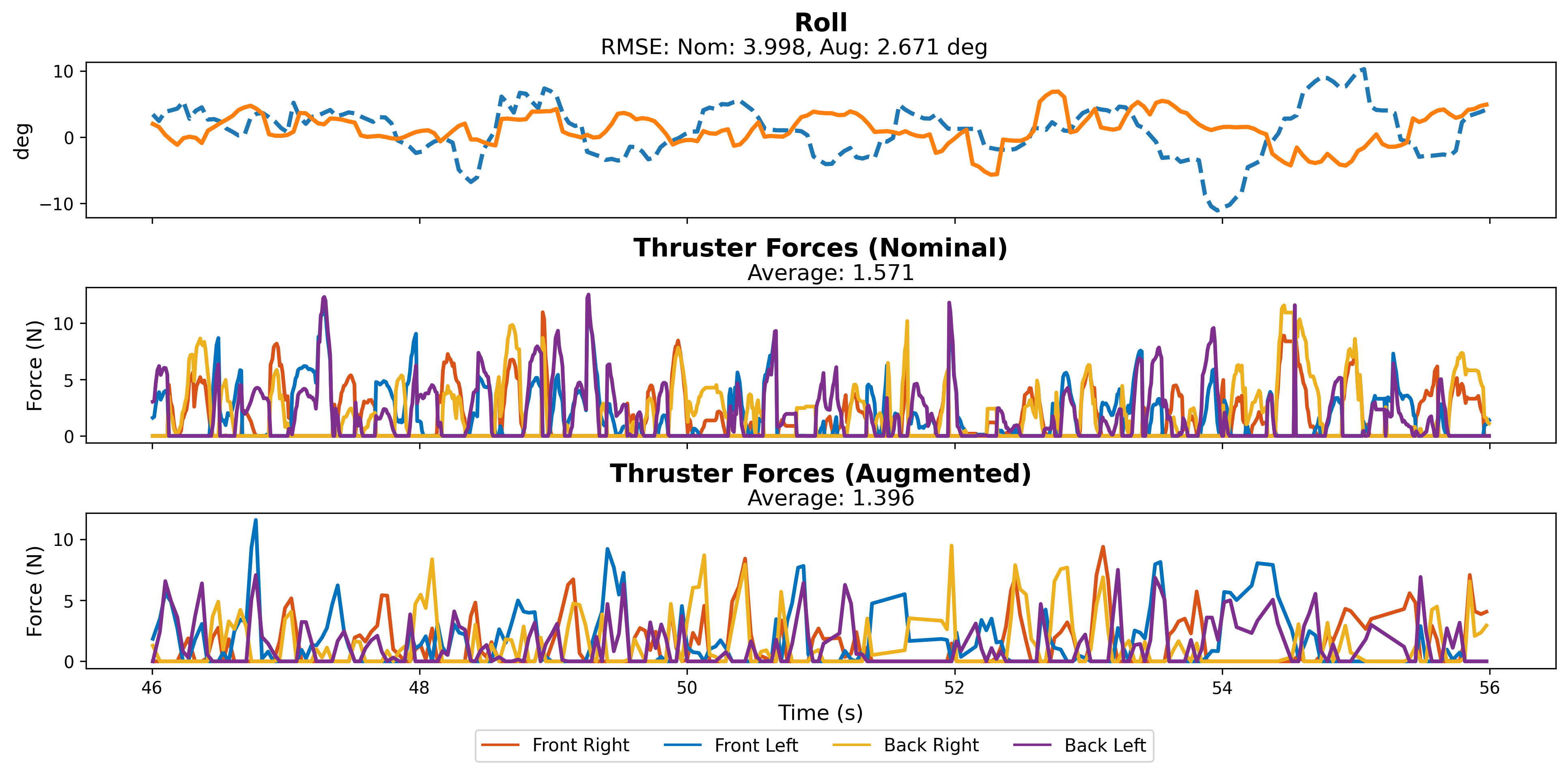}
    \caption{Roll stability and thruster force output comparison during normal-width gait trotting on hardware.}
    \label{fig:hw-result-normal-width}
\end{figure*}

In the final hardware experiment, normal-width gait trotting was evaluated (see Figure \ref{fig:hw-result-normal-width}). The CRD-augmented dynamic model not only achieved more stable roll dynamics but also required less thruster force compared to the nominal dynamic model, as its behavior was more closely aligned with the true dynamics.
\section{CONCLUSIONS}
We propose and implement a decoupled control architecture for thruster-assisted locomotion on Husky-$\beta$. For the legged part, a Raibert-type controller is utilized, while the ground impact generated by legged control is learned as a contact residual dynamic (CRD) within the thruster MPC formulation using a physics-informed loss function. Offline data are used to train the CRD network, which is then executed online. We further validate the proposed method in both simulation and hardware, where the augmented linear dynamic model more closely approximates the true dynamics, and finally, a cat-type walking gait is employed to demonstrate the effectiveness of the proposed control architecture in both simulation and hardware.

\FloatBarrier
\section{ACKNOWLEDGMENT}
This work is supported by the U.S. National Science Foundation (NSF) CAREER Award No. 2340278.

\addtolength{\textheight}{-12cm}   



\printbibliography

\end{document}